\documentclass[twoside]{article}

\usepackage[accepted]{aistats2025}
\usepackage[round]{natbib}

\usepackage[colorlinks=true,linkcolor=blue,citecolor=blue,urlcolor=blue]{hyperref}  

\usepackage{booktabs}
\usepackage{multirow}
\usepackage[table,xcdraw]{xcolor}
\usepackage{graphicx}

\begin{document}

%

%
\runningtitle{Application of SSMs  
to HEP with LSH}

\twocolumn[

\aistatstitle{Application of Structured State Space Models \\
to High energy physics with locality-sensitive hashing}

\aistatsauthor{ Cheng Jiang \And Sitian Qian }
\aistatsaddress{  University of Edinburgh \And  Peking University } ]

\begin{abstract}
  Modern high-energy physics (HEP) experiments are increasingly challenged by the vast size and complexity of their datasets, particularly regarding large-scale point cloud processing and long sequences. In this study, to address these challenges, we explore the application of structured state space models (SSMs), proposing one of the first trials to integrate local-sensitive hashing into either a hybrid or pure Mamba Model. Our results demonstrate that pure SSMs could serve as powerful backbones for HEP problems involving tasks for long sequence data with local inductive bias. By integrating locality-sensitive hashing into Mamba blocks, we achieve significant improvements over traditional backbones in key HEP tasks, surpassing them in inference speed and physics metrics while reducing computational overhead. In key tests, our approach demonstrated promising results, presenting a viable alternative to traditional transformer backbones by significantly reducing FLOPS while maintaining robust performance.
\end{abstract}

\section{INTRODUCTIONS}

Large-scale point clouds and long-sequence data processing are becoming prevalent in many scientific domains. In one of the fundamental subjects, high-energy physics (HEP), exploring the frontier of new physics usually requires a vast amount of data to support and validate the excess of theory prediction statistically \citep{higgs, CMS:2012qbp}. As the need for more data grows, the CERN Large Hadron Collider (LHC) \citep{HLLHC}, for example, must operate in increasingly intense experimental environments. Higher luminosity and collision rates are essential to probe deeper into the fundamental structure of matter, but this also leads to more complex data landscapes and more congested signatures of events, resulting in higher pile-up effects where multiple interactions occur within the same detector readout window. It is expected that a higher number of particles is directly related to increased complexity in tracking their trajectories and depositing energy across the detector in a finer granularity \citep{phase2}. 

The sequential nature of interactions over time, as particles traverse the detector, results in long sequence data. Each particle’s trajectory can be represented as a series of hits over time, and reconstructing these paths requires processing these long sequences of interactions. Traditional methods struggle with capturing long-range dependencies and missing intricate relationships between hits, which leads to information loss and reduced accuracy in the reconstruction, especially in an intensive environment. Similarly, these particle interactions can also be represented within a geometric space \citep{particlenet,pct,part,mpgan} and leverage geometric deep learning (GDL) \citep{gdl,gdl2} to model and analyze them effectively. Our study will primarily focus on the key challenges of tracking in High Luminosity-LHC \citep{HLLHC}, while also addressing the complexities of pileup mitigation. Both of these tasks involve processing data with very long, variable-length sequences. Furthermore, both tasks could benefit from introducing deep learning models with local inductive bias, which would help effectively associate nearby hits and energy deposits with the correct particle or interaction.

\begin{figure}[h]
    \centering  
    \includegraphics[width=0.98\linewidth]{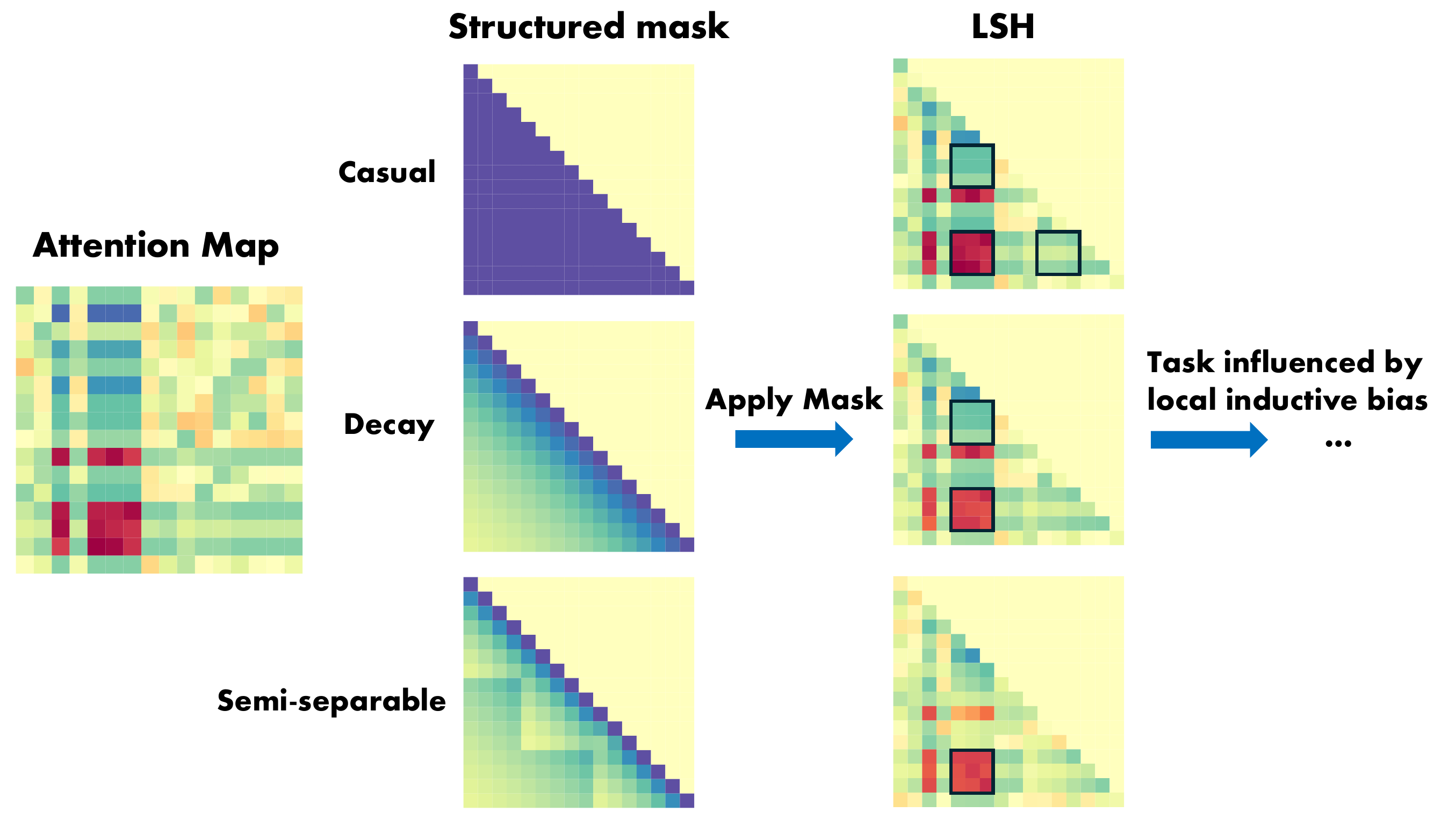}
    \caption{An illustration of different structured mask types applied to attention maps, followed by the integration of Local-Sentive Hashing (LSH). The causal mask corresponds to linear attention transformers, the decay mask to the retentive network, and the semi-separable mask represents structured SSM. In this context, LSH operates by aligning similar query-key values, allowing the semi-separable mask to effectively select the relevant blocks from neighboring regions while minimizing interference from unrelated areas.}
    \label{fig1}
\end{figure}

Previous studies have aimed to preserve the locality of point clouds by aggregating information from nearby regions, ensuring that local structures and interactions are accurately captured. A few recent approaches have utilized graph neural networks (GNNs) to dynamically construct graphs \citep{gnntrack,hgnn,deepcore,eggnet}, or have introduced specific objective functions like object condensation loss \citep{oc} to enhance the modeling of point clouds. However, those types of GNN applications have the intrinsic downside where traditional ways like k-NN or radius graph construction would introduce $\mathcal{O}(n^2)$ to make both training and inference very slow, especially when $n$ is larger. Transformers, on the other hand, are well-suited for capturing long-range dependencies and have demonstrated strong performance in various tasks. However, they also suffer from quadratic complexity due to the matrix manipulation required for full multi-head attention, which can become computationally expensive for large-scale point clouds or long sequences. Later trials aimed at reducing the complexity of the attention mechanism have explored approaches such as low-rank approximation \citep{lowrank}, retentive attention \citep{rententive}, and etc \citep{performer,reformer,linformer,smyrf}. Some studies \citep{hept} have shown the random Fourier features (RFF) \citep{rff} and locality-sensitive hashing (LSH) \citep{lsh} are so far few of the best efficient transformers to be used for tasks with local inductive bias. Despite these efforts, the computational overhead remains unsatisfactory, as the complexity still scales with $\mathcal{O}(n\text{poly}(n))$ \citep{hept}, which significantly hampers both training and inference in large-scale scenarios. None of the current methods have explored the usage of the State Space Model (SSM) \citep{ssm0,ssm1,ssm2,ssm3}, which offers near-linear complexity. SSMs are particularly well-suited for longer sequences and could comprehensively anatomize the real advantages of intensive tracking hits.

 Originally derived from the classic Kalman filter model, SSMs have proven highly efficient in processing long sequences due to their designated convolutional computation and low computational complexity. The early development of SSM did not give stronger metrics than traditional transformers in language modeling. Until very recently, models like Mamba \citep{s4,h3,mamba} and Griffin \citep{griffin} have shown comparable, if not superior, performance in a variety of modeling tasks involving long tokens. The first version of Mamba \citep{mamba} incorporates time-varying parameters into a recurrent dynamical system and utilizes a hardware-aware algorithm that enables highly efficient training and inference. The second version of Mamba \citep{mamba2} introduces State Space Duality, which establishes a connection between structured SSMs and variants of attention by interpreting SSMs as semi-separable matrix transformations. This approach eventually generalizes the way SSMs operate as structured masked linear attention. Some novel hybrid Transformer-Mamba architectures \citep{vim,jamba,zamba} were presented after this and showed even better performance than both pure Mamba and Transformer models.

The study is inspired to explore what might happen if one merges the efficient attention mechanism into Mamba, creating a hybrid model that could combine the strengths of both approaches. Additionally, we considered incorporating LSH into Mamba to determine whether this could further enhance performance in tasks that benefit from local inductive bias, such as tracking or other applications involving structured data. These modifications aim to balance computational efficiency with the ability to capture both local and long-range dependencies.

In this paper, we have conducted extensive validation on the proposed architectures, evaluating datasets ranging from 3k to over 60k tracking hits. Compared to the previous HEPT model, which already achieved state-of-the-art (SOTA) accuracy with more than 200$\times$ speedup, our approach further reduces FLOPs by over 10$\times$ times while maintaining comparable accuracy and significantly improving SOTA recall. Alternatively, we achieve SOTA accuracy with only a fraction of the model size. Additionally, in the pileup mitigation task, all proposed models outperform the previous studies. By offering both practical applicability and improved performance, those models could be much more realistic to apply in real HEP experiments.

\section{BACKGROUND}

\subsection{State Space Models}

Starting with the classical state space models that are continuous-time in nature, the new structured state-space sequence models try to use the knowledge from Recurrent Neural Network (RNN) \citep{rnn0,rnn1} and Convolutional Neural Network (CNN) \citep{cnn0,cnn1} to make the sequence model hold the ability to capture long-range dependencies at very low computational costs. The general form of structured SSMs can be written as:
\[
\begin{array}{ll}
h'(t) = Ah(t) + Bx(t)  & \quad y(t) = Ch(t) \quad \text{(1a)} \\
\quad h_t = \bar{A}h_{t-1} + \bar{B}x_t        & \quad y_t = Ch_t \quad \text{(1b)}
\end{array}
\]
The model is defined via four parameters($\Delta$, A, B, C) where the step size $\Delta$ controls how much focus is on the current state $x_t$. The first stage uses the discretization rule to transform “continuous parameters” $(\Delta, A, B)$ to “discrete parameters” $(\bar{A}, \bar{B})$ through fixed formulas \(\bar{A} = f_A(\Delta, A)\) and \(\bar{B} = f_B(\Delta, A, B)\). After the transformation, the model \citep{s4} can be computed by linear recurrence or global convolution.

Based on these concepts, Mamba introduces a refined mechanism to boost performance. The architecture begins by incorporating a standard local convolution before applying the traditional SSM steps, establishing the foundation of the Mamba architecture. This initial convolution allows the model to capture localized patterns and dependencies. Next, Mamba adopts a gated MLP structure, where the usual multiplicative gating mechanism is replaced with an activation function. Moreover, Mamba uses its selective copying mechanism, which addresses one of the key challenges faced by traditional transformer architectures. Unlike transformers \citep{transformer}, which rely on a Key-Value (KV) cache to store the entire sequence of content, Mamba can selectively choose which parts of the input to retain or ignore. The $1^{st}$ version of Mamba already showed better performance and exceptionally faster inference than the common transformer backbone in many language modeling tasks.

The $2^{nd}$ version of Mamba generalizes the structured state space model as the special form of attention family. Specifically, the linear attention uses the associativity of matrix multiplication in attention calculation \citep{linear} $\left( QK^\top \right) \cdot V = Q \cdot \left( K^\top V \right)$, and drops the softmax. If thinking of the casual (autoregressive) attention mask as a lower triangle matrix, then those efficient transformers can have a relatively general attention form $\left( L \circ QK^\top \right) \cdot V$ where mask matrix L determines how the final attention map looks like, as seen in Fig.~\ref{fig1}. By induction, we could write the hidden state $h_t$ as:
\[
\begin{array}{ll}
h_t &= A_t \cdots A_1 B_0 x_0 + A_t \cdots A_2 B_1 x_1 + \cdots + B_t x_t \\
    &= \sum_{s=0}^{t} A^{\times}_{t:s} B_s x_s. \quad \text{(2)}
\end{array}
\]
then $y_t$ in Eqn.1b is rewritten as:
\[
\begin{array}{ll}
y_t &= \sum_{s=0}^{t} C_t^T A^{\times}_{t:s} B_s x_s \quad \text{(3)}\\
y &= SSM(A, B, C)(x) = Mx \quad \text{(4)}\\
\end{array}
\]
The left matrix $M_{ji}:= C_j^T A_j \cdots A_{i+1} B_i$ is a semi-separable lower triangle matrix where every sub-rectangle block contained below the diagonal has rank up to SSM’s state dimension N. This generalization of a structured attention mask brings SSMs closer to attention mechanisms, making Mamba adaptable for both pure or hybrid models \citep{jamba,hydra}.

\subsection{Local Sensitive Hashing}

Locality-sensitive hashing (LSH) was initially introduced for approximate nearest neighbor search. It works by ensuring that, with high probability, data points that are close together are hashed into the same bucket, while points that are far apart are placed into different buckets. For E2LSH (Euclidean LSH) \citep{e2lsh}, which is designed for Euclidean distance, the hash function is defined as $H_{a,b}(x) = \left\lfloor \frac{a \cdot x + b}{r} \right\rfloor$ where $a$ is the unit normal distribution, $b$ is the random number uniformly distributed from 0 to $r$, $r > 0$ that controls bucket sizes. Two main logical LSH were introduced: OR\&AND LSH. In OR LSH, multiple hash tables are constructed, each using a different set of hash functions $g_1, g_2, \dots, g_{m_1}$. Two points are neighbors if they match in at least one of the hash tables, i.e., $g_i(x) = g_i(y)$ for some $i$. In AND LSH, multiple hash functions $H_1, H_2, \dots, H_{m_2}$ to form a new hash family $g(x) = [H_1(x), H_2(x), \dots, H_{m_2}(x)]$. Two points $x$ and $y$ are considered neighbors if all hash functions agree, i.e., $g(x) = g(y)$.

\begin{figure*}[t]
    \centering
    
    \includegraphics[width=0.7\textwidth]{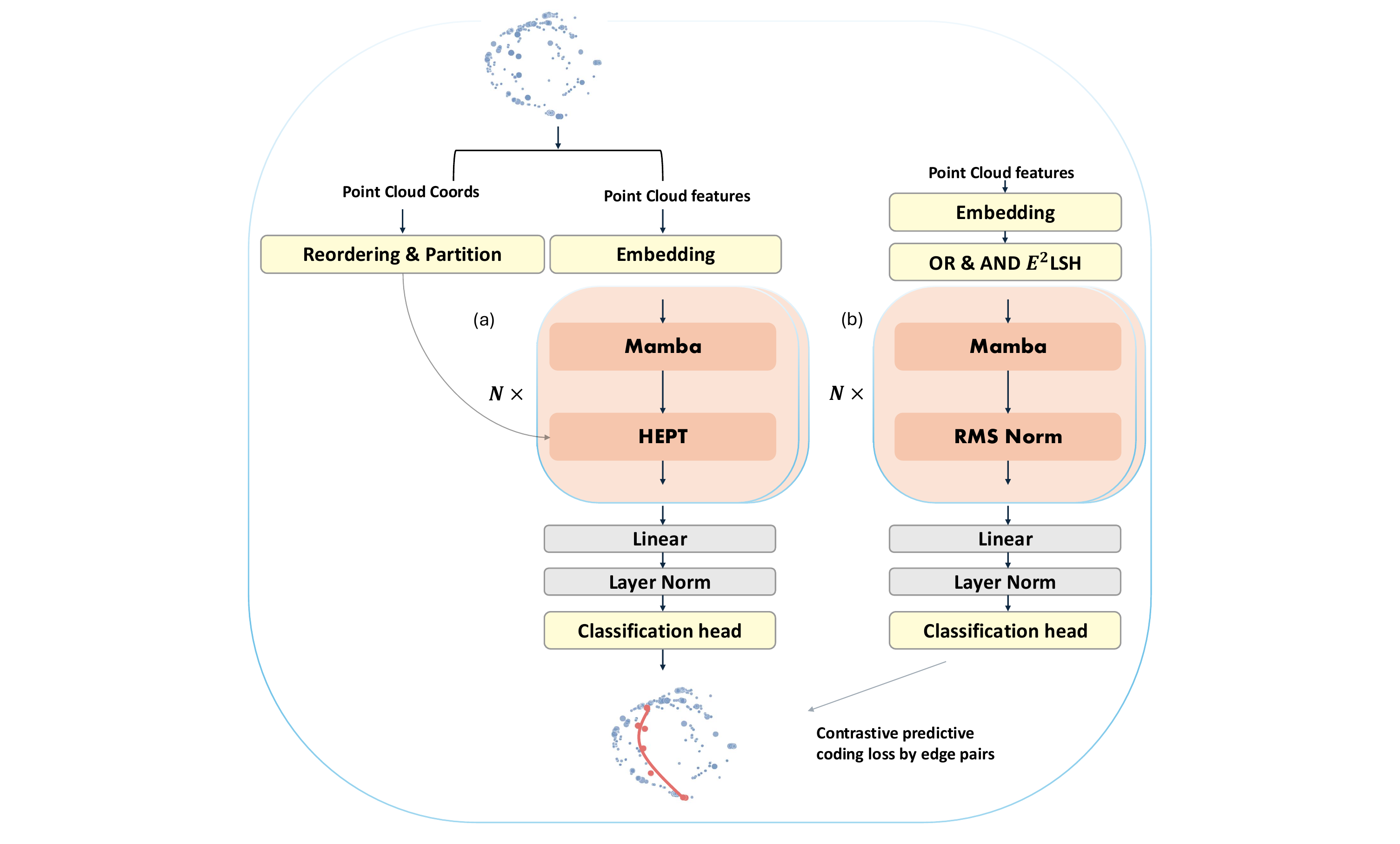}
    \caption{A schematic diagram of two proposed architectures for tasks with local inductive bias is presented. Left: The Mamba-a architecture, inspired by the hybrid Transformer-Mamba model (i.e., Jamba), excludes the MoE layer to reduce training memory requirements. Right: The Mamba-b architecture integrates the OR\&AND E2LSH selection and bucketing mechanism into pure Mamba blocks. The final loss for both models is computed from the embedding output using the Info Noise-Contrastive Estimation (InfoNCE) loss \citep{infonce}, constructed by predefined kNN edge pairs.}
    \label{fig2}
\end{figure*}

In most LSH-based efficient transformers like Reformer \citep{reformer} and Smyrf \citep{smyrf}, they adopt only the OR LSH and ignore the AND LSH. Since OR LSH only requires agreement on one of the hash tables rather than all possible hash functions, it can lead to a larger approximation error as the number of points increases. This relaxation reduces the precision of the approximation, making it more likely to group dissimilar points, especially in larger datasets. The recent work LSH-based Efficient Point Transformer (HEPT) tried to combine two hash tables and showed a good performance on tasks with local inductive bias. Similar to the query-key alignment in transformers, if hidden states $h_u$ and $h_v$ have small $||h_u-h_v||_2$, they are likely to have close hash values. We follow the same processing steps as HEPT, by further partitioning the selectively hidden space into various distinct, non-overlapping regions randomly to avoid large misalignments in the same bucket. More details can be found in \citep{hept}.

\subsection{Long Sequence Tasks in HEP}

\paragraph{Tracking} In HEP, tracking refers to the process of reconstructing the trajectories of charged particles produced in particle collision. Tracking is essential for understanding particle interactions, as it allows physicists to reconstruct the paths of particles based on measurements (called "hits") from detectors. These hits are associated with the positions where particles pass through the detector layers, and the tracking algorithm's goal is to associate them with the correct particle tracks, ideally with high accuracy and efficiency. \textbf{Traditional tracking algorithms at the LHC, surprisingly similar as SSM, are based on Kalman filters~\citep{BILLOIR1984352, FRUHWIRTH1987444}}. These algorithms work by iteratively extrapolating a particle’s trajectory based on a small set of initial hits (a "track seed") and searching for subsequent hits that match the predicted path. The computational cost of conventional algorithms increases rapidly as the number of hits and pileup events grows, which is particularly problematic for future HL-LHC, where tracking large numbers of particles in high-pileup environments becomes a computational bottleneck.

\paragraph{Pileup Mitigation}  Pileup refers to multiple particle collisions occurring simultaneously. Reconstructed objects falsely attributed to the primary collision are considered pileup contamination, which can inevitably degrade the accuracy and sensitivity of event reconstruction and analysis. Pileup mitigation techniques are designed to identify and remove particles resulting from pileup interactions, ensuring that only particles from the primary collision are included in the analysis. One of the widely-used algorithms for pileup mitigation is charged-hadron subtraction, as implemented in the CMS Particle Flow (PF) algorithm \citep{chf}. This method rejects charged particles whose tracks are not associated with the primary collision vertex, thus reducing pileup contamination. In more advanced approaches, algorithms such as PUPPI \citep{puppi} (PileUp Per Particle Identification) and SoftKiller \citep{softkiller} aim to use likelihood-based method to suppress pileup contributions among neutral particles as well.

\section{RELATED WORKS}

\paragraph{LSH-based Efficient Point Transformer} As the paper mentioned earlier, the HEPT architecture leverages both OR LSH and AND LSH to approximate the attention mechanism with near-linear computational complexity. The model starts by evaluating the error-computation tradeoff to argue the LSH-based efficient transformers have smaller approximation errors than RFF-based ones. To address potential misalignment issues between query and key buckets, HEPT integrates point cloud coordinates as additional AND hash codes. This ensures that only queries and keys that are close in both feature space and geometric space are grouped, which further reduces the approximation error. The performance validation of the model gives SOTA accuracy on long sequence tracking while still maintaining the same scale of inference FLOPs as other common efficient transformers. 

\paragraph{Hybrid SSM-attention model} A few hybrid SSM-attention models, such as Zamba, use shared attention and MLP block to connect multiple Mamba blocks with shared parameters. This setup leverages the FLOPs efficiency of Mamba blocks while delivering performance comparable to other language models. Another approach to constructing hybrid models involves combining Transformer, Mamba, and MoE across different layers. However, in our empirical tests on those two tasks, the inclusion of MoE did not consistently yield a clear performance boost, while significantly increasing RAM consumption. So, we decided to drop the MoE setting when introducing the hybrid model Mamba-a, as shown in Fig.~\ref{fig2}. 

\paragraph{Receptance Weighted Key Value} Receptance Weighted Key Value (RWKV) \citep{rwkv} is a novel model architecture designed to combine the efficient inference of RNNs with the parallelizable training of transformers. One key addition to the QKV mechanism in transformers is the Receptance vector, which controls how much information from the previous time step should be carried forward, similar to the gating mechanism in RNNs. It also has a weighted decay vector which modifies the interaction weights between time steps. The basic RWKV architecture consists of two main components: time-mixing and channel-mixing. The time-mixing part works similarly to the attention mechanism, while the channel-mixing part resembles the feed-forward layer. In RWKV5, the model incorporates the concept of multi-head computation. RWKV6 further improves the architecture by integrating Low-Rank Adaptation (LoRA) \citep{lora} into the weight decay factor, making the model more efficient in terms of fine-tuning. In this study, we will use RWKV as the validation set for our proposed Mamba architecture, as it also claims to have near-linear computational complexity. Yet we believe there is significant potential for further exploration of RWKV in future research.

\begin{figure*}[t]
    \centering
    \begin{minipage}[t]{0.48\textwidth}
        \centering
        \includegraphics[width=\textwidth]{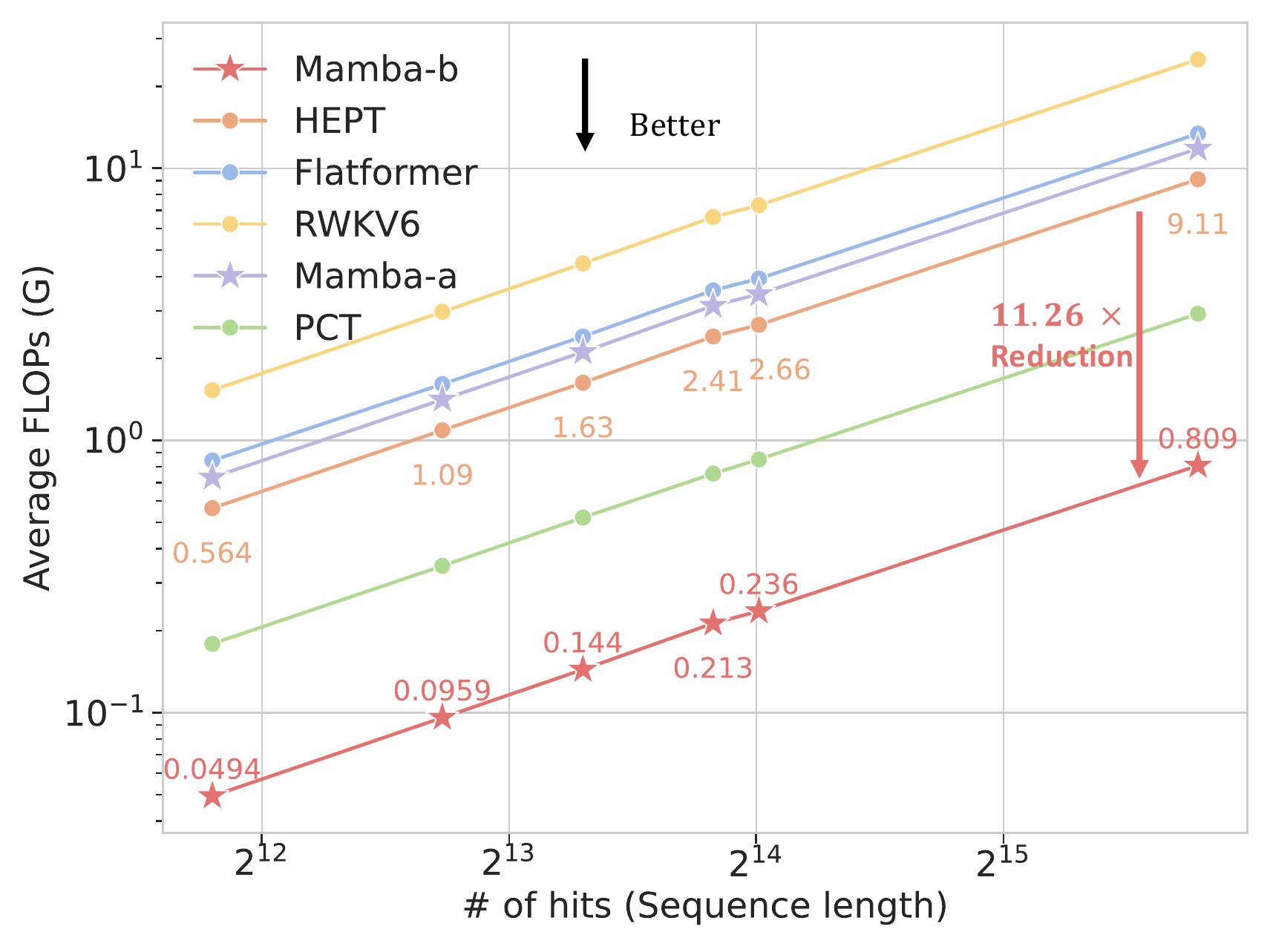}
        \caption{Performance plots for average inference FLOPs of various small-sized models across different numbers of hits, ranging from 3k to 60k. (All tests were performed on the actual dataset and evaluated on a single NVIDIA A100, ensuring realistic performance evaluation rather than relying on toy points.)}
        \label{fig3}
    \end{minipage}
    \hfill
    \begin{minipage}[t]{0.48\textwidth}
        \centering
        \includegraphics[width=\textwidth]{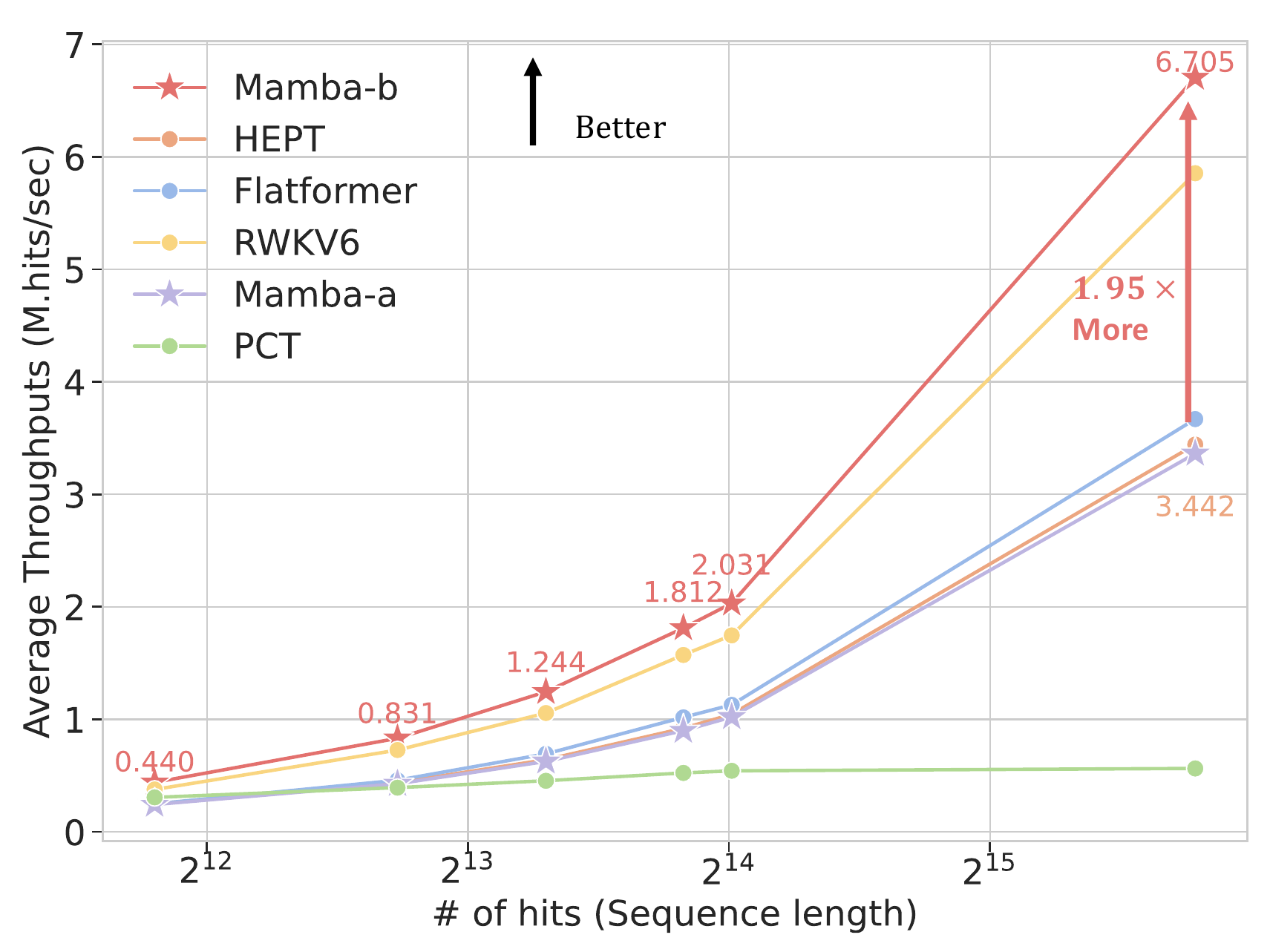}
        \caption{Performance plots for average throughputs (millions of hits) of various small-sized models across different numbers of hits, ranging from 3k to 60k. (while the actual FLOPs do not represent fully the raw inference time, can see clearly from the PCT and Flatformer case.)}
        \label{fig4}
    \end{minipage}
\end{figure*}
\section{TRAINING METHOD}

\subsection{Datasets}
The tracking dataset is used in the TrackML challenge \citep{trackml1,trackml2}, simulating detector conditions under the HL-LHC condition. Each event contains simulated 3D measurements (hits) of particles generated during proton collisions at LHC. The objective is to group the recorded hits into particle tracks, where each hit should be associated with a single particle track. Each hit is given by the unique hit ID, 3D coordinates, and geometry-related identifiers. We consider only particles with transverse momentum threshold pT $>$ 0.9 GeV as the low pT event usually would come from background noise. One complete event has more than 60k hits, we can continue splitting the events by different numbers of sectors.\footnote{ Note the different number of sectors would not equally split the hits in complicated geometry, in the final performance plots Fig.~\ref{fig3},~\ref{fig4}, we choose $n_{sector}$ to be 1,2,3,6,10,20 to test the performance under different lengths.}

For the pileup mitigation tasks, we used a publicly available dataset \citep{pu} simulated under the same HL-LHC pileup conditions. Each event in the dataset has particle-level information such as four-momentum, charge, and other relevant properties. The dataset consists of 1,000 point clouds, each containing approximately 10,000 points.

\subsection{Training Setup}

For the tracking dataset, we convert the 3D Euclidean coordinates into cylindrical coordinates and utilize both local and global coordinates of the sub-detector. For more details, refer to Table.6 in \citep{hept}. All low-level inputs are embedded into hidden dimensions before being processed by any blocks. The Mamba-a architecture, as shown on the left of Fig.~\ref{fig2}, applies the same HEPT attention layer after every Mamba layer. The same reordering and partitioning strategy is employed for LSH. After passing through the main blocks, the data is fed into a fully connected layer without bias and undergoes layer normalization. Finally, a classification MLP is applied to produce the output in the desired embedding dimension. The Mamba-b architecture, as shown on the right of the figure, only applies the OR\&AND LSH once before being processed by the main blocks. Essentially, the Mamba-b architecture should have roughly the same model complexity as the pure Mamba model, with the only addition being the LSH processing applied before the main block to minimize the overhead. 

For the pileup dataset, we have a total of 7 particle-level features, including information such as location, momentum, and energy. The Particle ID, which indicates the type of particle, is embedded into an 8-16 dimensional space before being combined with the positional embedding. Later training setup follows the same approach as the tracking dataset, but the model outputs predictions on a per-particle basis.

\section{EXPERIMENTAL RESULTS}

The evaluation metrics for tracking consider two key aspects. First, inference speed is crucial, as fast machine learning inference becomes increasingly important as the number of data points grows larger in real-experiment scenarios. We measured the average inference FLOPs and throughput for each tracking event across the entire dataset, with hit counts ranging from 3k to 60k.  To fully understand the actual computation effectiveness, it is also essential to consider the raw inference time, as FLOPs alone do not fully stand for the actual inference speed. From the performance plot Fig.~\ref{fig3},~\ref{fig4}, we used a single NVIDIA A100 GPU to calculate the FLOPs during the forward pass. Combining the results from both figures, the Mamba-a-S model demonstrates FLOPs and throughput on par with other RFF-based models like FlatFormer \citep{flatformer} and LSH-based transformers like HEPT. Although PCT has significantly fewer FLOPs compared to other models (except for Mamba-b-S), it exhibits worse scalability and higher computational complexity as the number of hits increases. Conversely, RWKV shows the highest FLOPs but also the second-highest throughput. Most notably, Mamba-b-S achieves both the best performance in terms of FLOPs, with an 11.26× reduction compared to the previous SOTA model HEPT, and the highest throughput, boasting 1.96× more than the usual efficient transformers, all while maintaining a similar model size.

\begin{figure}[h]
    \centering
    \includegraphics[width=\linewidth]{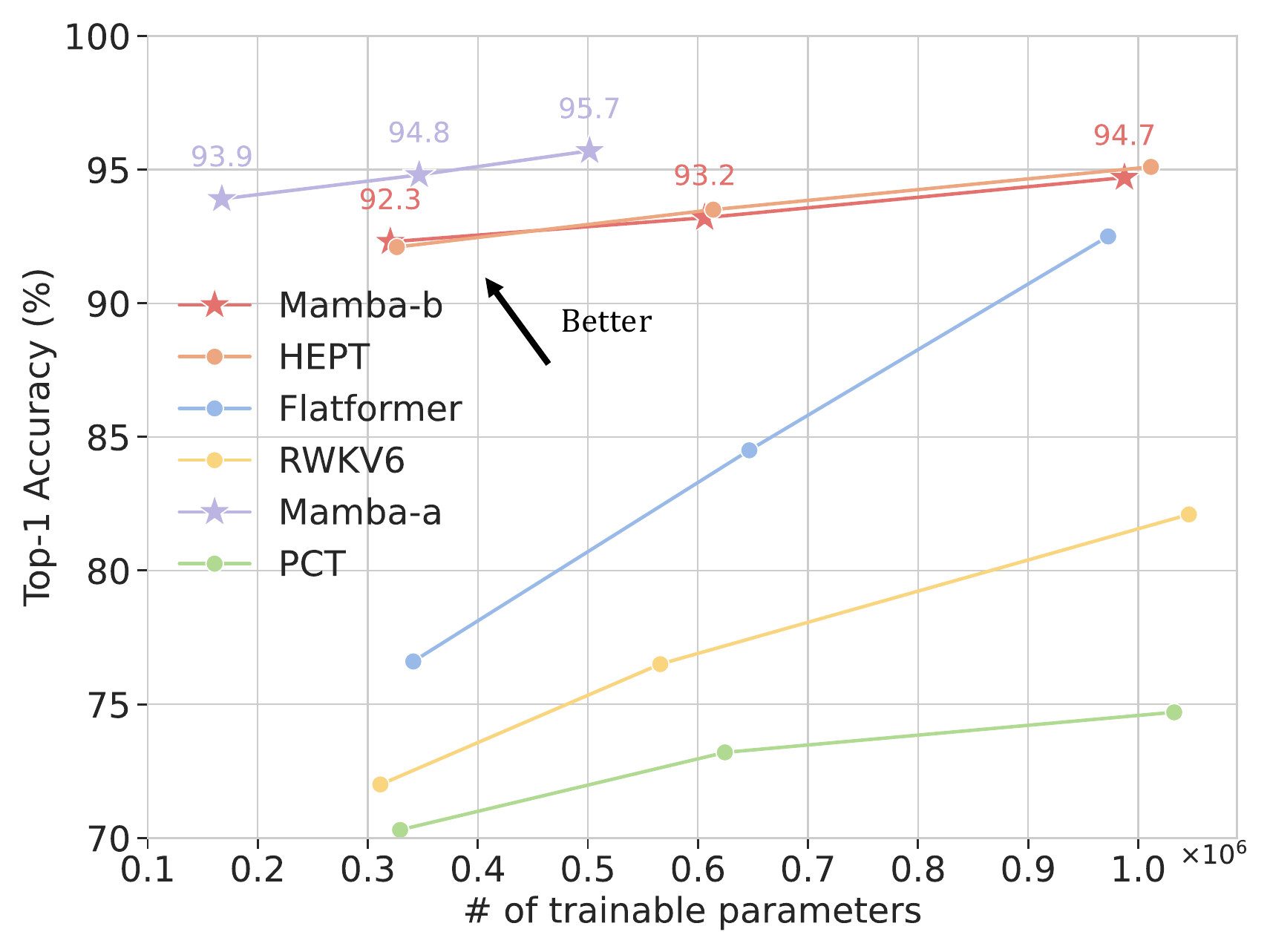}
    \caption{Top-1 accuracy of different-sized models on Tracking-60k scenario. Mamba-a demonstrates better accuracy than all other larger models, while Mamba-b achieves comparable accuracy as the previous SOTA model.}
    \label{fig5}
\end{figure}

\begin{table}[htp]
\caption{Top1-Accuracy table for different backbones with different numbers of parameters on Tracking-60k dataset. The \textasteriskcentered, \textdagger, \textdaggerdbl, are the first, second, and third best results. \textbf{Bold} is the highlighted result with great performance.} 
\label{table1}
\hspace{5mm}
\centering
\renewcommand{\arraystretch}{1.2} 
\arrayrulecolor{black} 
\begin{tabular}{|p{0.3\linewidth}|p{0.2\linewidth}|>{\centering\arraybackslash}p{0.3\linewidth}|}
\toprule
\textbf{Method} & \textbf{\#param.} & \textbf{Tracking-60k top-1 acc.} \\
\specialrule{0.1em}{0.2em}{0.2em} 
\multicolumn{3}{c}{\textbf{RNN}} \\
\specialrule{0.05em}{0.1em}{0.1em} 
RWKV5-S      & 0.32 M  & 72.4 \\
\specialrule{0.05em}{0.1em}{0.1em}
RWKV5-M      & 0.57 M  & 76.6 \\
\specialrule{0.05em}{0.1em}{0.1em}
RWKV6-S      & 0.31 M  & 72.0 \\
\specialrule{0.05em}{0.1em}{0.1em}
RWKV6-M    & 0.57 M  & 76.5 \\
\specialrule{0.05em}{0.1em}{0.1em}
RWKV6-L & 1.05 M  & 82.1 \\

\specialrule{0.1em}{0.2em}{0.2em} 
\multicolumn{3}{c}{\textbf{Transformers}} \\
\specialrule{0.05em}{0.1em}{0.1em}
HEPT-S         & 0.33 M  & 92.1\textsuperscript{\textdaggerdbl} \\
\specialrule{0.05em}{0.1em}{0.1em}
HEPT-M        & 0.61 M & 93.5\textsuperscript{\textdagger} \\
\specialrule{0.05em}{0.1em}{0.1em}
HEPT-L      & 1.01 M   & \textbf{95.1}\textsuperscript{\textdagger} \\
\specialrule{0.05em}{0.1em}{0.1em}
FlatFormer-S     & 0.34 M  & 76.6 \\
\specialrule{0.05em}{0.1em}{0.1em}
FlatFormer-M       & 0.65 M  & 82.5 \\
\specialrule{0.05em}{0.1em}{0.1em}
FlatFormer-L       & 0.97 M  & 91.0 \\
\specialrule{0.05em}{0.1em}{0.1em}
PCT-S     & 0.33 M  & 70.3 \\
\specialrule{0.05em}{0.1em}{0.1em}
PCT-M       & 0.61 M  & 73.2 \\
\specialrule{0.05em}{0.1em}{0.1em}
PCT-L       & 1.03 M  & 74.7 \\
\specialrule{0.1em}{0.2em}{0.2em} 
\multicolumn{3}{c}{\textbf{SSMs}} \\
\specialrule{0.05em}{0.1em}{0.1em}
Mamba-S    & 0.33 M  & 79.4 \\
\specialrule{0.05em}{0.1em}{0.1em}
Mamba-M    & 0.63 M  & 80.9 \\
\specialrule{0.05em}{0.1em}{0.1em}
Mamba-L    & 1.00 M  & 84.4 \\
\specialrule{0.05em}{0.1em}{0.1em}
\rowcolor[gray]{0.9} Mamba-a-S   & 0.17 M   & 93.9\textsuperscript{\textasteriskcentered} \\
\specialrule{0.05em}{0.1em}{0.1em}
\rowcolor[gray]{0.9} Mamba-a-M   & 0.35 M   & 94.8\textsuperscript{\textasteriskcentered} \\
\specialrule{0.05em}{0.1em}{0.1em}
\rowcolor[gray]{0.9} Mamba-a-L   & 0.50 M   & \textbf{95.7}\textsuperscript{\textasteriskcentered} \\
\specialrule{0.05em}{0.1em}{0.1em}
\rowcolor[gray]{0.9} Mamba-b-S   & 0.32 M   & 92.3\textsuperscript{\textdagger} \\
\specialrule{0.05em}{0.1em}{0.1em}
\rowcolor[gray]{0.9} Mamba-b-M   & 0.61 M   & 93.2\textsuperscript{\textdaggerdbl} \\
\specialrule{0.05em}{0.1em}{0.1em}
\rowcolor[gray]{0.9} Mamba-b-L   & 0.99 M   & \textbf{94.7}\textsuperscript{\textdaggerdbl} \\
\bottomrule
\end{tabular}

\end{table}

\begin{table}[h]
\caption{Best top backbones with small numbers of parameters on Tracking-6k (short), Tracking-15k (medium), Tracking-60k (long) dataset. \textbf{Bold} is the highlighted result with great performance.} 
\label{table2}
\hspace{5mm} 
\centering
\renewcommand{\arraystretch}{1.2} 
\arrayrulecolor{black} 
\begin{tabular}{|p{0.3\linewidth}|>{\centering\arraybackslash}p{0.25\linewidth}|>{\centering\arraybackslash}p{0.25\linewidth}|}
\toprule
\textbf{Method} & \textbf{top-1 acc.} & \textbf{top-1 recall} \\
\specialrule{0.1em}{0.2em}{0.2em} 
\multicolumn{3}{c}{\textbf{Tracking-6k}} \\
\specialrule{0.1em}{0.2em}{0.2em} 
HEPT-S         & 92.2  & 95.6 \\
FlatFormer-S         & 77.1  & 95.2 \\
\rowcolor[gray]{0.9}Mamba-a-S         & \textbf{94.0}  & 96.5 \\
\rowcolor[gray]{0.9}Mamba-b-S         & 91.9  & \textbf{99.1}  \\

\multicolumn{3}{c}{\textbf{Tracking-15k}} \\
\specialrule{0.1em}{0.2em}{0.2em} 
HEPT-S         & 92.2  & 95.7 \\
FlatFormer-S         & 76.7  & 94.2 \\
\rowcolor[gray]{0.9}Mamba-a-S         & \textbf{94.3}  & 96.9\\
\rowcolor[gray]{0.9}Mamba-b-S         & 91.5  & \textbf{99.0} \\

\multicolumn{3}{c}{\textbf{Tracking-60k}} \\
\specialrule{0.1em}{0.2em}{0.2em} 
HEPT-S         & 92.1  & 95.7 \\
FlatFormer-S         & 76.6  & 94.0 \\
\rowcolor[gray]{0.9}Mamba-a-S         & \textbf{93.9}  & 96.7 \\
\rowcolor[gray]{0.9}Mamba-b-S         & 92.3  & \textbf{99.3} \\
\bottomrule
\end{tabular}
\end{table}

\begin{figure}[h]
    \centering
    \includegraphics[width=\linewidth]{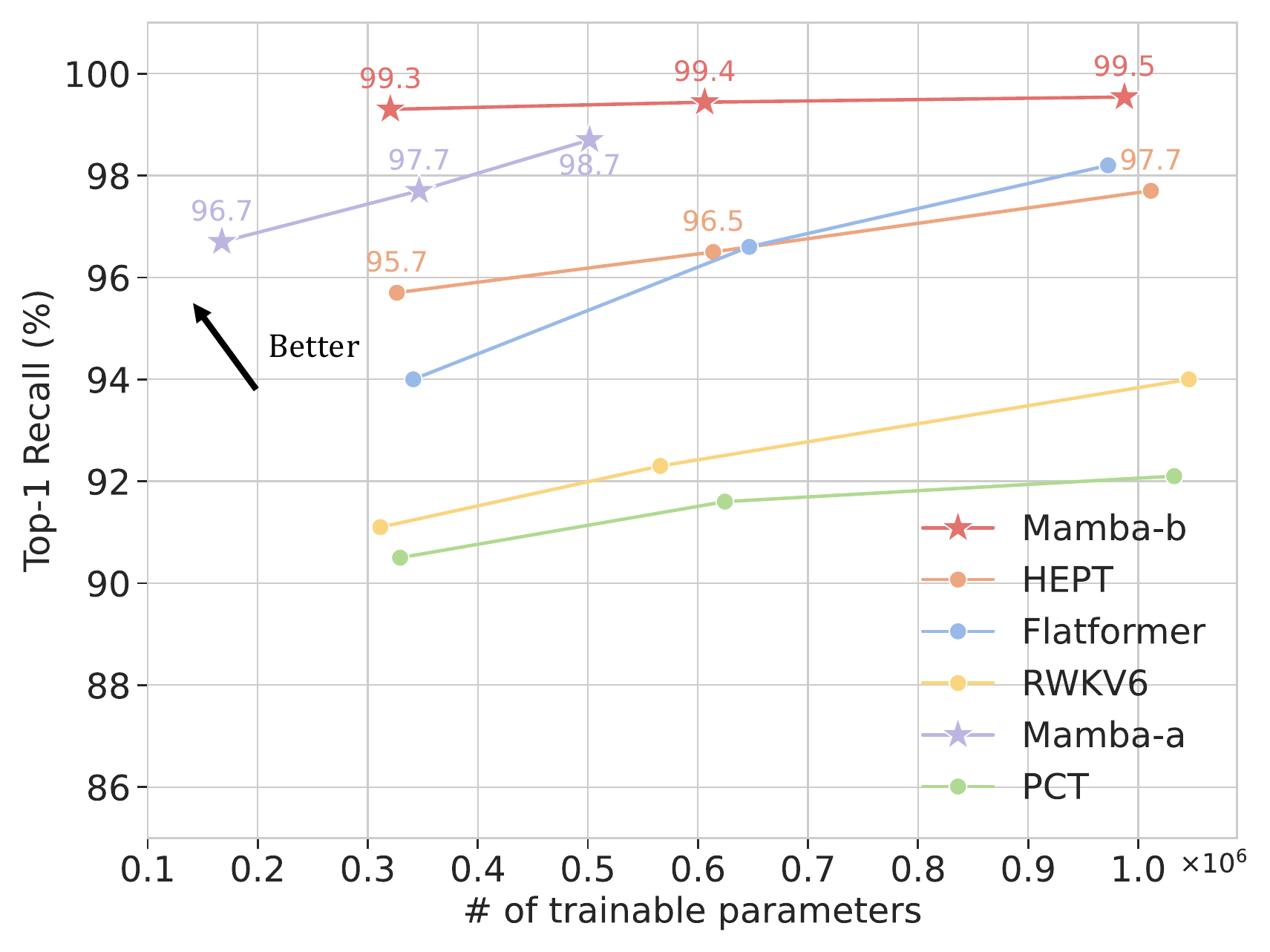}
    \caption{Top-1 recall of different-sized models on Tracking-60k scenario. Mamba-b demonstrates significantly better recall than all other larger models.}
    \label{fig6}
\end{figure}

\begin{table}[htp]
\caption{Summary of top-1 accuracy and the top-1 area under curve score for ROC. \textbf{Bold} for best performance.}  
\label{table3}
\hspace{5mm}
\centering
\renewcommand{\arraystretch}{1.2} 
\arrayrulecolor{black} 
\begin{tabular}{|p{0.3\linewidth}|>{\centering\arraybackslash}p{0.25\linewidth}|>{\centering\arraybackslash}p{0.25\linewidth}|}
\toprule
\textbf{Method} & \textbf{top-1 acc.} & \textbf{top-1 ROC} \\
\specialrule{0.1em}{0.2em}{0.2em} 
\multicolumn{3}{c}{\textbf{Pileup-10k}} \\
\specialrule{0.1em}{0.2em}{0.2em} 
HEPT-S         & 40.35 & 75.56 \\
FlatFormer-S         & 38.87  & 75.05 \\
PCT-S              & 40.40  & 75.91 \\
\rowcolor[gray]{0.9} Mamba-a-S         & 40.39  & 75.77 \\
\rowcolor[gray]{0.9} Mamba-b-S         & \textbf{40.43} & \textbf{76.01} \\
\bottomrule
\end{tabular}
\end{table}

Second, the accuracy and recall of the model are equally important. In addition to general performance metrics, we use a specialized metric to assess the quality of the learned point embeddings, particularly how well the embeddings from the same particle cluster are grouped. acc. is defined as $\frac{1}{n} \sum_{u=1}^{n} \text{Prec@}k_u$ where the hits originating from the same particle as hit $u$ (i.e. by checking if the neighbor labels match the true labels, as precision). Recall is defined as $\frac{1}{n} \sum_{u=1}^{N} \text{Prec@}k_u$ where the summation includes all the matched clusters in one event. Ideally, the recall value should closely resemble the perfect match efficiency in a physical case. This would mean that the number of reconstructed tracks includes all hits from the matched particle and excludes any other hits \footnote{In LHC experiments, tracking efficiency requires that at least 75\% of the hits in a track belong to the same particle. This criterion typically results in higher efficiency compared to the measured recall here}.

The final evaluation metrics for different model sizes in Tracking-60k are presented in Fig.\ref{fig5}, Fig.\ref{fig6}, and Table~\ref{table1}. From previously studied models, we observed that performance could still be improved, particularly for efficient transformers. HEPT achieved up to 95.1\% accuracy with around 1 million parameters. The most significant improvement was seen in FlatFormer, as its performance increased notably with a higher number of parameters. The Mamba-a architecture, consistently achieves the highest accuracy with fewer trainable parameters, already surpassing the best performance of HEPT with only about one-third of the model size. While measuring recall, HEPT or models that use hybrid HEPT attention typically result in lower recall at the same accuracy level. We found that the Mamba-b architecture, which relies solely on the pure Mamba backbone, achieves significantly higher recall, nearly approaching 1. 

We have summarized our results in Table~\ref{table2}, comparing different models across varying numbers of hits. The best four models were selected for comparison. Mamba-a-S consistently delivers the highest accuracy across all datasets, while Mamba-b-S shows a notably higher recall than any other models for all datasets.

For pileup mitigation, we emphasize performance in the binary classification tasks. In Table~\ref{table3}, the best performance is achieved by the Mamba-b architecture. Additionally, we empirically found that small-sized PCT models can be trained to achieve performance comparable to or even better than HEPT models and the hybrid Mamba-PCT model can therefore also deliver excellent performance. 


\section{LIMITATIONS}

While we have demonstrated significant improvements compared to previous models, there are still a few limitations to be addressed. 

Since the tasks in this study primarily focus on local inductive bias, a future direction could involve exploring the model's robustness across a broader range of HEP tasks, where different patterns or dependencies may arise. As emphasized, different tasks might have different preferences toward the trade-off between computing throughput and physics performance.

We used a public dataset that contains limited statistics for pileup mitigation. In a more ideal scenario, having access to more detailed statistics could enable more comprehensive physical evaluations as done for the tracking experiment.


\section{CONCLUSION}
In this study, we presented one of the first attempts to bring SSMs into HEP tasks. Additionally, we introduced the use of OR \& AND E2LSH before applying Mamba's selective mechanism to integrate LSH-based attention into a hybrid Transformer-Mamba architecture, which is designed to tackle tasks with local inductive bias. This approach addresses two key challenges in HEP: HL-LHC tracking and pileup mitigation. Our quantitative metrics show that the proposed architecture reduces FLOPs by over 10$\times$ compared to SOTA models with recall significantly improved and accuracy remaining comparable or even better.

\subsubsection*{Code Availability}

Our code is available at \url{https://github.com/chengjiang123/Lampa}.

\subsubsection*{Acknowledgements}
We thank Dr. Huilin Qu, Dr. Yihui Lai, and Dr. Yongbin Feng for the fruitful discussions.

\bibliographystyle{plainnat}
\bibliography{ref}

\section*{Checklist}

 \begin{enumerate}

 \item For all models and algorithms presented, check if you include:
 \begin{enumerate}
   \item A clear description of the mathematical setting, assumptions, algorithm, and/or model. [Yes]
   \item An analysis of the properties and complexity (time, space, sample size) of any algorithm. [Yes]
   \item (Optional) Anonymized source code, with specification of all dependencies, including external libraries. [Yes]
 \end{enumerate}

 \item For any theoretical claim, check if you include:
 \begin{enumerate}
   \item Statements of the full set of assumptions of all theoretical results. [Not Applicable]
   \item Complete proofs of all theoretical results. [Not Applicable]
   \item Clear explanations of any assumptions. [Not Applicable]
 \end{enumerate}

 \item For all figures and tables that present empirical results, check if you include:
 \begin{enumerate}
   \item The code, data, and instructions needed to reproduce the main experimental results (either in the supplemental material or as a URL). [Yes]
   \item All the training details (e.g., data splits, hyperparameters, how they were chosen). [Yes]
         \item A clear definition of the specific measure or statistics and error bars (e.g., with respect to the random seed after running experiments multiple times). [Yes]
         \item A description of the computing infrastructure used. (e.g., type of GPUs, internal cluster, or cloud provider). [Yes]
 \end{enumerate}

 \item If you are using existing assets (e.g., code, data, models) or curating/releasing new assets, check if you include:
 \begin{enumerate}
   \item Citations of the creator If your work uses existing assets. [Yes]
   \item The license information of the assets, if applicable. [Yes]
   \item New assets either in the supplemental material or as a URL, if applicable. [Not Applicable]
   \item Information about consent from data providers/curators. [Yes]
   \item Discussion of sensible content if applicable, e.g., personally identifiable information or offensive content. [Not Applicable]
 \end{enumerate}

 \item If you used crowdsourcing or conducted research with human subjects, check if you include:
 \begin{enumerate}
   \item The full text of instructions given to participants and screenshots. [Not Applicable]
   \item Descriptions of potential participant risks, with links to Institutional Review Board (IRB) approvals if applicable. [Not Applicable]
   \item The estimated hourly wage paid to participants and the total amount spent on participant compensation. [Not Applicable]
 \end{enumerate}

 \end{enumerate}

\setcounter{section}{0}

%

%



\renewcommand{\thesection}{\Alph{section}}


%

%

\onecolumn
\aistatstitle{Supplementary material}

\section{Details on experiment setup}

\subsection{Loss function choices}

In tracking tasks, we improved the original loss function from \citep{hept}, where they adopted contrastive learning with InfoNCE loss \citep{infonce}.

\begin{align}
    \mathcal{L}_{\mathrm{InfoNCE}} = - \log \frac{\exp(\mathrm{sim}(h_u, h_v^+))}{\exp(\mathrm{sim}(h_u, h_v^+)) + \sum_{h_v^- \in \mathcal{N}} \exp(\mathrm{sim}(h_u, h_v^-))}
\end{align}

Pairs of hits from the same particle are labeled positive ($\{h^+\}$), while negative  ($\{h^-\}$) pairs are selected from the nearest 256 hits ($\mathcal{N}$) belonging to other particles. The $L^2_\mathrm{RBF}$ similarity 
\begin{align}
    \mathrm{sim}_{{L^2_\mathrm{RBF}}}(h_u, h_v) = \exp\left(-\frac{d_{uv}^2}{2 \sigma^2}\right) 
    , \hspace{5mm} d_{uv} = \| h_u - h_v \|_2 = \sqrt{\sum_{k=1}^{\mathcal{N}} (h_{u,k} - h_{v,k})^2} 
\end{align}
is used.  Originally, neighboring hits were pre-defined by constructing a radius graph, but this approach significantly increased the required training RAM and data size. To address this, we refine the neighbors with a kNN ($\mathcal{N} \to k$) leveraging the inductive bias on locality. Similar performance could be achieved with significantly reduced memory consumption by setting $k$ to a small value (e.g., 32).

In pileup tasks, each sample event contains an unequal number of charged and neutral particles. To address this challenge, we use Focal Loss \citep{focal}, 
\begin{align}
    \mathcal{L}_{\mathrm{F.L.}} = -\alpha_t(1-p_t)^\lambda \log(p_t)
\end{align}
with $\lambda$ set to 2, $\alpha$ set to 0.25 for class imbalance.

\newpage
\subsection{Model Hyperparameters}


For the tracking task, it is crucial to maintain high throughput while pushing for better performance. Therefore, we tested our model with different scales to study the scaling behavior.

We chose to set up our experiment with three different scales of our model, denoted as Small (S), Medium (M), and Large (L). 

The Small (S) model is designed to have a similar size, matching the 0.33M parameters of HEPT chosen in previous literature. The Medium (M) model doubles the size of the Small model, while the Large (L) model is scaled up to have approximately 1 million trainable parameters. This range of model sizes allows us to analyze the trade-offs between model complexity and performance across different tasks. 

To scale up the models, two main variables, the embedding dimension of the hidden state and the number of layers, are varied. The embedding dimension of the hidden state is set to either 24 or 48. The number of layers is set to 4, 8, and 12. For the LSH hyperparameters, we choose the number of hashes (m1, m2) to be fixed at 3. The block size (Table~\ref{tablea}) varies depending on the number of tracking hits and the chosen model, namely Mamba-a or Mamba-b.

\begin{table}[h!]
\centering
\caption{Block size used in different datasets}

\vspace{5mm}
\centering
\renewcommand{\arraystretch}{1.2} 
\arrayrulecolor{black} 
\begin{tabular}{|p{0.2\linewidth}|>{\centering\arraybackslash}p{0.22\linewidth}|>{\centering\arraybackslash}p{0.22\linewidth}|}
\toprule
\textbf{Dataset}  & \textbf{Mamba-a} & \textbf{Mamba-b} \\ 
\specialrule{0.1em}{0.2em}{0.2em}

\textbf{Tracking-6k}  & [20, 40]                 & [100, 120]                  \\ 
\textbf{Tracking-15k} & [60, 80, 100]            & [150, 200, 250]             \\ 
\textbf{Tracking-60k} & [140, 150, 160]          & [200, 250, 300]             \\
\textbf{Pileup-10k}   & [100, 120, 140]          & [100, 150, 200]             \\ 
\bottomrule
\end{tabular}
\label{tablea}
\end{table}

For the pile-up tasks, the performance is more crucial, hence we have one fixed hyperparameter configuration. The embedding dimension and number of layers are set to {24, 8} for Mamba-a, {48, 8} for Mamba-b, LSH hyperparamters are shown on the last row in Table.~\ref{tablea}


\end{document}